\documentclass{article}
\usepackage{ijcai22}

\usepackage{times}
\usepackage{soul}
\usepackage{url}
\usepackage{verbatim}
\usepackage[hidelinks]{hyperref}
\usepackage[utf8]{inputenc}
\usepackage[small]{caption}
\usepackage{graphicx}
\usepackage{amsmath}
\usepackage{amsthm}
\usepackage{booktabs}
\usepackage{algorithm}
\usepackage{algorithmic}
\usepackage{pifont}
\newcommand{\cmark}{\ding{51}}%
\usepackage{multirow}
\usepackage{bbm}
\usepackage{comment}
\usepackage{caption}
\usepackage{subcaption}

\title{Region-level Contrastive and Consistency Learning for Semi-Supervised Semantic Segmentation}

\author{
Jianrong Zhang$^1$\textsuperscript{\thanks{Equal contribution}\thanks{Interns at the Institute of Deep Learning, Baidu Research}}
\and
Tianyi Wu$^{2,3}$ \textsuperscript{\footnotemark[1]}
\and
Chuanghao Ding$^{4}$ \textsuperscript{\footnotemark[1]}
\and
Hongwei Zhao$^1$ 
\and
Guodong Guo$^{2,3}$ \textsuperscript{\thanks{Corresponding author}}
\affiliations
$^1$College of Computer Science and Technology, Jilin University, Changchun, China\\
$^2$Institute of Deep Learning, Baidu Research, Beijing, China\\
$^3$National Engineering Laboratory for Deep Learning Technology and Application, Beijing, China\\
$^4$College of Software, Jilin University, Changchun, China\\
\emails
\{jrzhang20, dingch20\}@mails.jlu.edu.cn,
zhaohw@jlu.edu.cn, \\
\{wutianyi01, guoguodong01\}@baidu.com
}

\begin{document}

\maketitle

\begin{abstract}
Current semi-supervised semantic segmentation methods mainly focus on designing pixel-level consistency and contrastive regularization. However, pixel-level regularization is sensitive to noise from pixels with incorrect predictions, and pixel-level contrastive regularization has memory and computational cost with $O(pixel\_num^2)$. To address the issues, we propose a novel region-level contrastive and consistency learning framework (RC$^2$L) for semi-supervised semantic segmentation. Specifically, we first propose a Region Mask Contrastive (RMC) loss and a Region Feature Contrastive (RFC) loss to accomplish region-level contrastive property. Furthermore, Region Class Consistency (RCC) loss and Semantic Mask Consistency (SMC) loss are proposed for achieving region-level consistency. Based on the proposed region-level contrastive and consistency regularization, we develop a region-level contrastive and consistency learning framework (RC$^2$L) for semi-supervised semantic segmentation, and evaluate our RC$^2$L on two challenging benchmarks (PASCAL VOC 2012 and Cityscapes), outperforming the state-of-the-art.
\end{abstract}

\section{Introduction}

Semantic segmentation has high potential values in a variety of applications. 
However, training supervised semantic segmentation models requires large-scale pixel-level annotations,
and such pixel-wise labeling is time-consuming and expensive. This work focuses on semi-supervised semantic segmentation, which takes advantage of a large amount of unlabeled data and limits the need for labeled examples. 

\begin{figure}[t]
    \centering
    \includegraphics[scale=0.72]{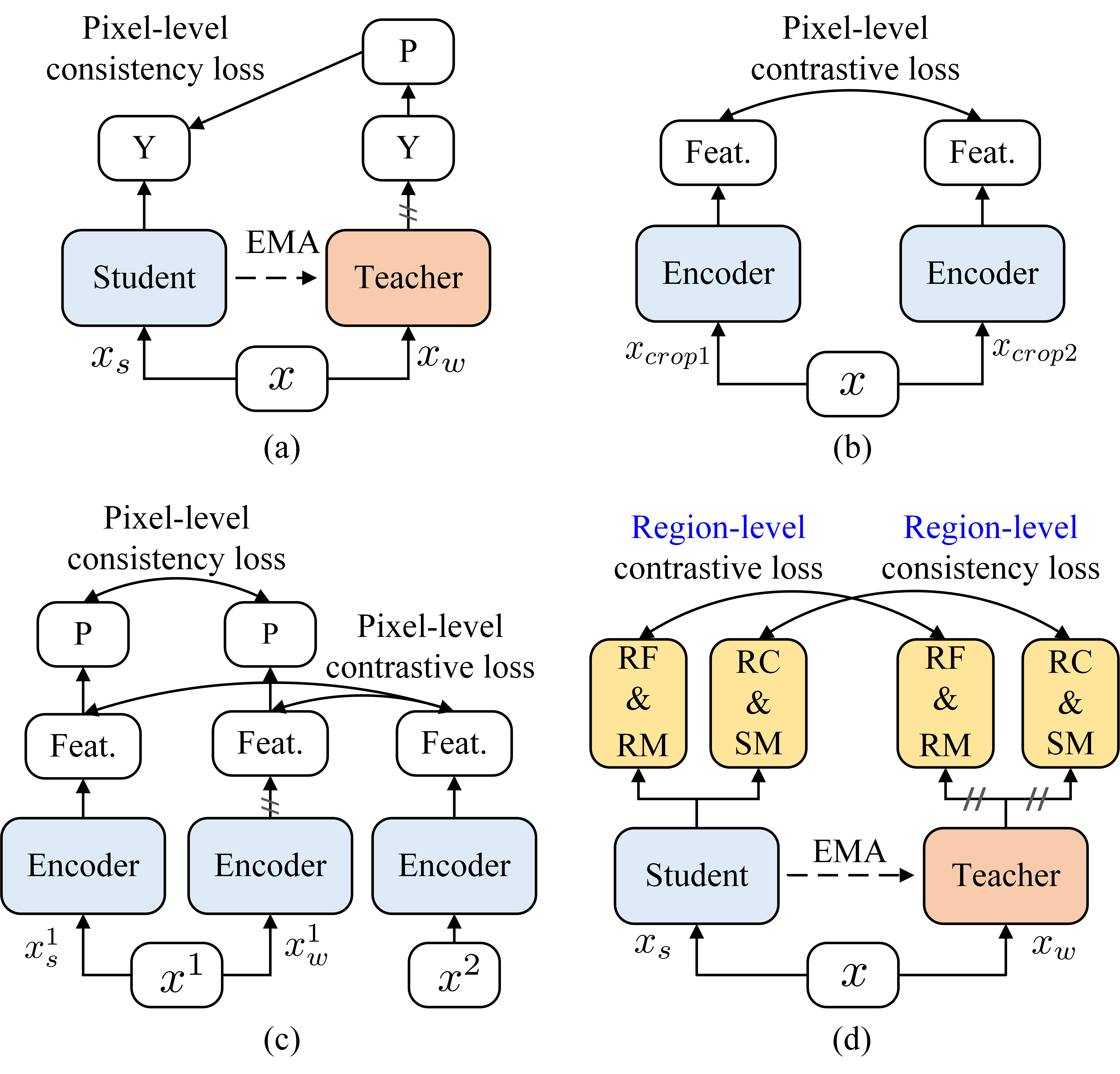}
    \caption{Comparison with existing semi-supervised semantic segmentation methods. (a) Pixel-level label consistency~\protect\cite{french2020:cutmixseg}, (b) Pixel-level feature contrastive property~\protect\cite{lai2021semi}, (c) Combination of pixel-level label consistency and feature contrastive property~\protect\cite{zhong2021:pixel_contrastive}. (d) Region-level contrastive property and consistency (Ours). `` // '' on $\rightarrow$ means stop-gradient. P: pseudo labels, Y: predicted probability maps, RF: region features, RM: region masks, RC: region classes, SM: semantic masks.}
    \label{fig1}
\end{figure}

Currently, many works have demonstrated that designing pixel-level consistency regularization \cite{french2020:cutmixseg,ouali2020CCT,chen2021:CPS},
and pixel-level contrastive regularization \cite{zhong2021:pixel_contrastive,lai2021semi}  are beneficial for semi-supervised semantic segmentation. 
The first family of works utilized pixel-level label consistency under 
different data augmentations \cite{french2020:cutmixseg,zou2021pseudoseg},
feature perturbations \cite{ouali2020CCT}, network branches \cite{GCT}, and segmentation models \cite{chen2021:CPS}.
These methods benefited from using the label-space consistency property on the unlabeled images.
Figure \ref{fig1} (a) shows the pixel-level label consistency of employing different data augmentations. 
Differently, the work \cite{lai2021semi} proposed the Directional Contrastive Loss to accomplish 
the pixel-level feature contrast, which is shown in Figure \ref{fig1} (b).
Substantial progress has been made by utilizing pixel-level label consistency and pixel-level feature contrastive property.
For example, PC$^2$Seg~\cite{zhong2021:pixel_contrastive} leveraged and simultaneously enforced the consistency property in the label space and the contrastive property in the feature space, which is illustrated in Figure \ref{fig1} (c).
However, such pixel-level regularization is sensitive to noise from pixels with incorrect predictions. 
Besides,  pixel-level contrastive regularization has memory and computational cost with $O(pixel\_num^2)$, and usually requires designing additional negative and positive example filtering mechanism carefully.

To overcome the above challenges, we propose to design Region-level Contrastive and Consistency Learning (RC$^2$L) for semi-supervised semantic segmentation. As shown in Figure \ref{fig1} (d), our method enforces the consistency of region classes and the contrastive property of features and masks from different regions. It was inspired by 
MaskFormer \cite{cheng2021:maskformer} which formulated supervised semantic segmentation as a mask (or region) classification problem.

Specifically, we first propose a Region Mask Contrastive (RMC) loss and a Region Feature Contrastive (RFC) loss to achieve region-level contrastive property. The former pulls the masks of matched region pairs (or positive pairs) closer and pushes away the unmatched region pairs (or negative pairs), and the latter pulls the features of matched region pairs closer and pushes away the unmatched regions pairs.
Furthermore, Region Class Consistency (RCC) loss and Semantic Mask Consistency (SMC) loss are proposed for encouraging the consistency of the region classes and the consistency of union regions with the same class, respectively.

Based on the proposed components, we develop a Region-level Contrastive and Consistency Learning (RC$^2$L) framework for semi-supervised semantic segmentation. The effectiveness of our approach is demonstrated by conducting extensive ablation studies.
In addition, we evaluate our RC$^2$L on several widely-used benchmarks, e.g., PASCAL VOC 2012 and Cityscapes, 
and the experiment results show that our approach outperforms the state-of-the-art semi-supervised segmentation methods. 

\section{Related works}

\paragraph{Semantic segmentation.}Since the emergence of Fully Convolutional Network (FCN)~\cite{long2015FCN}, per-pixel classifications have achieved high accuracies in semantic segmentation tasks. The development of modern deep learning methods for semantic segmentation mainly focuses on how to model context~\cite{chen2018deeplab,wu2020cgnet,wu2020ginet}. Recently, Transformer-based methods for semantic segmentation show an excellent performance.
SegFormer~\cite{xie2021segformer} proposed mask transformer as a decoder. 
More recently, MaskFormer~\cite{cheng2021:maskformer} reformulated semantic segmentation as a mask (or region) classification task. Differently, we focus on how to make a better use of unlabeled data to perform region-level predictions for semi-supervised semantic segmentation.

\paragraph{Semi-supervised semantic segmentation.}It is important to explore semi-supervised semantic segmentation to reduce per-pixel labeling costs. Early approach~\cite{hung2018adversarial} used generative adversarial networks (GANs) and adversarial loss to train on unlabeled data.
Consistency regularization and pseudo-labeling have also been widely explored for semi-supervised semantic segmentation, by enforcing consistency among the predictions, either from feature perturbation~\cite{ouali2020CCT}, augmented input images~\cite{french2020:cutmixseg}, or different segmentation models~\cite{GCT}.
Later, PseudoSeg~\cite{zou2021pseudoseg} combined pixel-level labels with image-level labels to enhance 
the semi-supervised learning through boosting the quality of pseudo labels. CPS~\cite{chen2021:CPS} proposed the cross pesudo supervision to apply consistency between two segmentation networks with the same architecture. 
Self-training has also driven the development of the state-of-the-art methods~\cite{rethink,he2021-re-dis}. 
Pixel-level contrastive based approaches have been proposed recently. CMB~\cite{alonso2021cmb} introduced a memory bank for positive-only contrastive learning. 
Directional Contrastive Loss~\cite{lai2021semi} was proposed for training between pixel features and pseudo labels. PC$^2$Seg~\cite{zhong2021:pixel_contrastive} enforced the pseudo labels to be consistent, and encouraged pixel-level contrast to pixel features.
All these methods mainly focused on pixel-level regularization and conducted pixel-level predictions. 
Differently, our approach explores region-level regularization and performs region-level predictions.

\paragraph{Contrastive learning.}Image-level contrastive learning has shown excellent prospects for self-supervised representation learning. SimCLR~\cite{chen2020:simclr} proposed to take the contrastive loss for training between images by applying different data augmentations. MoCo V2~\cite{chen2020mocov2} presented a momentum encoder to reduce the requirement of large batch size. Pixel-level contrastive learning has also been proven to be beneficial for dense prediction tasks. ~\cite{Contrastive-seg} introduced pixel-level contrastive learning for supervised semantic segmentation. Different from these works, we explore how to conduct region-level contrastive learning for semi-supervised semantic segmentation.

\begin{figure*}[ht]
\centering
\includegraphics[scale=1.0]{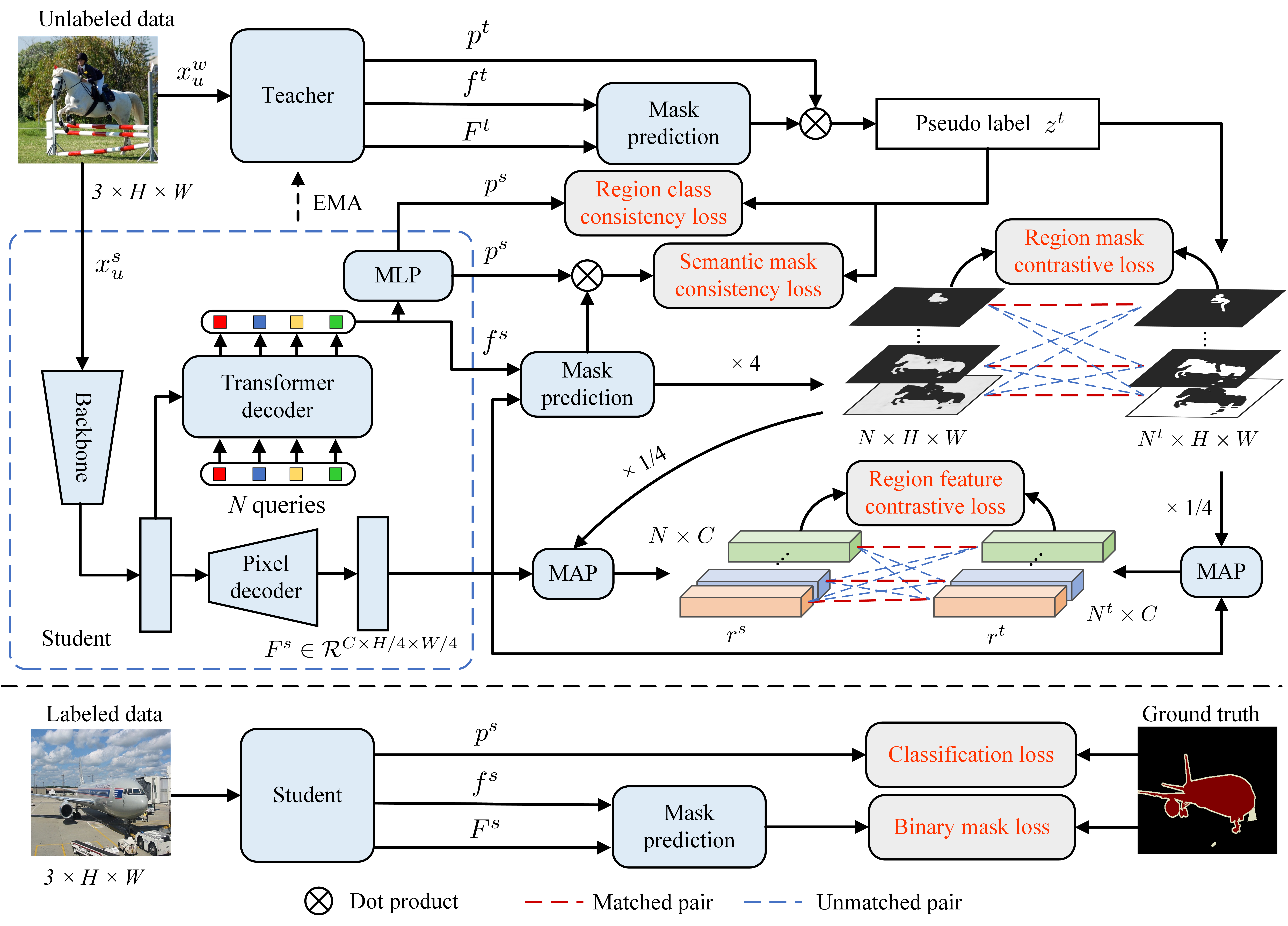}
\caption{Illustrate the architecture of our proposed Region-level Contrastive and Consistency. $p^s$ (or $p^t$) and  $f^s$ (or $f^t$) denote mask logits and mask embeddings of the student model (or teacher model), respectively. $F^s$ (or $F^t$) is the pixel decoder output feature map of the student model (or teacher model). $r^s$ (or $r^t$) represents region features of the student model (or teacher model). $H$ and $W$ are the height and width of the input image.  ``MAP'' denotes the mask average pooling. $N$ denotes the number of queries in an image from student model, and $N^t$ denotes the number of masks in an image from teacher model, respectively. $C$ is the embedding dimension.}
\label{fig2}
\end{figure*}

\section{Method}

We first describe our framework, Region-level Contrastive and Consistency Learning (RC$^2$L) for semi-supervised semantic segmentation. Then, we present the Region Mask Contrastive (RMC) loss and Region Feature Contrastive (RFC) loss.
The former pulls the masks of matched region pairs closer and pushes away the unmatched region pairs, 
and the latter pulls the features of matched region pairs closer and pushes away the unmatched region pairs. 
Finally, we introduce the Region Class Consistency (RCC) loss and Semantic Mask Consistency (SMC) loss. 
RCC loss can enforce the consistency of region classes, and SMC loss can further promote the consistency of union regions with the same class.

\subsection{Framework}
\label{sec30}
The overall framework of our RC$^2$L is shown in Figure \ref{fig2}, which consists of a student model and a teacher model. The teacher model has the same architecture as the student model but uses a different set of weights. The student model is trained with both labeled (with ground-truth) and unlabeled data.

Specifically, given an image-label pair, $\{x_l,z^{gt}\}$, $x_l$ is the image, and $z^{gt}$ is the corresponding annotations that are the set of $N^{gt}$ ground truth segments, i.e., $z^{gt} =\{ (c_i^{gt}, m_i^{gt}) | c_i^{gt} \in \{1,2,...,K\}, m_i^{gt} \in \{0,1\}^{H \times W}  \}_{i=1}^{N^{gt}}$, where $c_i$ is the ground truth class of the $i^{th}$ segment, $W$ and $H$ represent the width and height of the input image. We feed the image $x_l$ into the student model, outputting probability-mask pairs $z^l=\{ (p_i^l, m_i^l) | p_i^l \in \triangle^{K+1}, m_i^l \in [0,1]^{H \times W}  \}_{i=1}^{N}$, where the probability distribution $p_i$ contains a ``no object" label $\oslash$ and $K$ category labels.
A bipartite matching-based assignment $\sigma$ between the set of predictions $z^l$ and $z^{gt}$ is conducted for computing supervised loss:
\begin{equation}
    \begin{split}
        \mathcal{L}^{label}(z^l,z^{gt}) = & \sum_{i=1}^{N} [ -log p_{\sigma(i)}^l(c_i^{gt}) + \\
        & 1_{c_i^{gt} \neq \oslash }\mathcal{L}_{mask}(m_{\sigma (i)}^l, m_{i}^{gt})],
    \end{split}
\end{equation}
where $\mathcal{L}_{mask}$ is a binary mask loss. Please refer to \cite{cheng2021:maskformer} for more details.

For the unlabeled image $x_u$, we feed its weak augmentation $x_u^w$ and strong augmentation $x_u^s$ into the teacher and student models, respectively. Here, we use short edge resize, random crop, random flip, and color augmentation as weak augmentations, and use all weak data augmentation methods and CutMix \cite{yun2019cutmix} as strong augmentations.
The teacher model is employed to generate pseudo label $z^t = \{ (c_i^{t}, m_i^{t}) \}_{i=1}^{N^t}$ which is used to guide the training of the student model. The unsupervised loss can be formulated as follow:
\begin{equation}
    \begin{split}
        \mathcal{L}^{unlabel} (x_u) = \beta_1 \mathcal{L}_{RCC} + \beta_2 \mathcal{L}_{SMC} + \\
        \beta_3 \mathcal{L}_{RMC} + \beta_4 \mathcal{L}_{RFC} ,
    \end{split}
\end{equation}
where $\mathcal{L}_{RCC} $ and $\mathcal{L}_{SMC}$ mean the Region Class Consistency (RCC) loss and Semantic Mask Consistency (SMC) loss (Section \ref{sec31}), respectively. $\mathcal{L}_{RMC} $ and $\mathcal{L}_{RFC}$ denote the Region Mask Contrastive (RMC) loss and Region Feature Contrastive (RFC) loss (Section \ref{sec32}), respectively. $\beta_1$, $\beta_2$, $\beta_3$ and $\beta_4$ denote the loss weight. Therefore, the total loss is defined as:
\begin{equation}
\mathcal{L} = \mathcal{L}^{label}(z^l,z^{gt}) + \alpha \mathcal{L}^{unlabel}(x_{u}),
\end{equation}
where $\alpha$ is a constant to balance between the supervised and unsupervised losses.

Following Mean Teacher~\cite{tarvainen2017:MT}, the parameters  $\theta_t$ of teacher model are an  exponential moving average of parameters  $\theta_s$  of the student model. Specifically, at every training step, the parameters $\theta_t$ of teacher network is updated as follows:
\begin{equation}
\theta_t = \tau \theta_t + (1-\tau) \theta_s,
\end{equation}
where $\tau \in \ [0,1 \ ] $ is a decay rate.

\subsection{Region-level Contrastive Learning}
\label{sec31}
The goal of region-level contrastive learning is to increase the similarity between each region mask and feature from strong augmentation $x_u^s$ and the mask and feature of the matched region (from weak augmentation $x_u^w$), 
and reduce the similarity between unmatched region pairs.
Specifically, we propose a Region Mask Contrastive (RMC) loss and a Region Feature Contrastive (RFC) loss to achieve region-level contrastive property. The former pulls the masks of matched region pairs (or positive pairs) closer 
and pushes away the unmatched region pairs (or negative pairs), and the latter pulls the features of matched 
region pairs closer and pushes away the unmatched regions pairs.

The weak augmentation $x_u^w$ is firstly fed into the teacher network, which outputs pseudo labels $z^t = \{ (c_i^{t}, m_i^{t}) \}_{i=1}^{N^t}$.
The strong augmentation $x_u^s$ is fed into the student network, which outputs class logits $p^s=\{p_i^s \in \triangle^{K+1} \}_{i=1}^{N}$, 
mask embeddings $f^s = \{f_i^s \in \mathcal{R}^{C} \}_{i=1}^{N}$ , and per-pixel features $F^s \in \mathcal{R}^{C \times H/4 \times W/4}$. 
Firstly, we obtain binary mask predictions $ \{ m_i^{s} \}_{i=1}^{N}$ via a dot product between the mask embeddings and per-pixel features. 
Then, the bipartite matching-based assignment $\sigma$ between student predictions $\{ m_i^{s} \}_{i=1}^{N}$ and pseudo segment set $\{m_i^{t}\}_{i=1}^{N^t}$ is used for getting matched index set $ID = \{\sigma(i)\}_{i=1}^{N^t}$ and computing RMC Loss:
\begin{equation}
\mathcal{L}_{RMC} = \sum_{i}^{N^t} -log \frac{exp^{d(m_{\sigma(i)}^s, m_{i}^{t})/\tau_m}}{\sum_{\{j \in ID,    j\neq \sigma(i)\}}^N exp^{d(m_j^s, m_i^t)/\tau_m}},
\end{equation}
where $\tau_m$ is a temperature hyper-parameter to control the scale
of terms inside exponential, and $d(m_i^s,m_j^t)=\frac{2| m_i^s \cap m_j^t | }{ | m_i^s | \cup | m_j^t | }$ measures the similarity between two masks. 

Next, we compute region features $r^s =\{r_{\sigma(i)}^s\}_{i=1}^{N^t}$ via combining per-pixel features $F^s$ and region mask set $\{ m_{i}^{s} \}_{i=1}^{N}$. The process can be formulated as follow:
\begin{equation}
r_{\sigma(i)}^s = GAP( m_{\sigma(i)}^{s} \cdot F^s), \forall \sigma(i),
\end{equation}
where ``GAP'' indicates a global average pooling operation. Similarly, we compute target region features  $r^t =\{r_i^t\}_{i=1}^{N^t}$ as follow:
\begin{equation}
r_i^t = GAP( m_i^{t} \cdot F^s), \forall i,
\end{equation}
Here, per-pixel features $F^t$ are not used to compute the target region features, since the student model and the teacher model have different weights and feature space. Then, Region Feature Contrastive loss is defined as:
\begin{equation}
\mathcal{L}_{RFC} = \sum_{i}^{N^t} -log \frac{exp^{cos(r_{\sigma(i)}^s, r_{i}^{t})/\tau_f}}{\sum_{\{j \in ID,    j\neq \sigma(i)\}}^N exp^{cos(r_j^s, r_i^t)/\tau_f}},
\end{equation}
where $\tau_f$ is a temperature hyper-parameter to control the scale
of terms inside exponential, $cos(u, v) = \frac{u^\mathrm{ T } v}{\lVert u \rVert \lVert v \rVert} $ is the cosine similarity.

\paragraph{Comparison with Pixel Contrastive Loss.}The proposed Region Mask Contrastive (RMC) loss and Region Feature Contrastive (RFC) loss are different from the most related Pixel Contrastive Loss~\cite{zhong2021:pixel_contrastive} in two aspects. Firstly, it performed contrastive learning on pixel-level features, while our method conducts contrastive learning on region-level masks and features. Secondly, unlike \cite{zhong2021:pixel_contrastive}, our methods do not require designing the negative example filtering strategy, since there is no overlap for different region masks within the same image.

\subsection{Region-level Consistency Learning}
\label{sec32}
Different from previous works \cite{zhong2021:pixel_contrastive,alonso2021cmb} which used pixel-level consistency regularization, we develop a region-level consistency learning, which consists of a Region Class Consistency (RCC) Loss and Semantic Mask Consistency (SMC) loss. The former enforces the consistency of region classes, and the latter further promotes the consistency of the union regions with the same class.

Given the student class logits $\{p_i^s \in \triangle^{K+1} \}_{i=1}^{N}$ and the pseudo segment label $ \{c_i^{t}, m_i^{t} \}_{i=1}^{N^t}$, the proposed Region Class Consistency loss is employed to enforce the class consistency of matched region pairs, which is defined as:
\begin{equation}
\mathcal{L}_{RCC} = \sum_{i}^{N^t}  -log \, p_{\sigma(i)}^{s} (c_i^{t}),
\end{equation}

Furthermore, different regions may correspond to the same class, 
so we design a Semantic Mask Consistency loss to promote the consistency of the union regions with the same class and the pseudo semantic mask, which can be formulated as follow:
\begin{equation}
\mathcal{L}_{SMC}  = \sum_{i}^{N^t} \mathcal{L}_{mask} ( Union(\{m_i^s\}_{i=1}^{N}, c_i^t), m_i^t),
\end{equation}
where $Union(,)$ means merging regions with the same class into a single region. Following DETR~\cite{DETR} and MaskFormer~\cite{cheng2021:maskformer}, $\mathcal{L}_{mask}$ is a linear combination of a focal loss \cite{lin2017focal} and a dice loss \cite{milletari2016fully}.

\begin{table*}[ht]\small
    \centering\setlength{\tabcolsep}{3.5pt}
    \begin{tabular}{l|c|c|c|c|c|c|c|c|c}
        \toprule
        \multirow{2}*{Method} & \multicolumn{5}{c|}{VOC \texttt{Train}} & \multicolumn{4}{c|}{VOC \texttt{Aug}} \\ \cmidrule{2-10}
         & \small{1/2(732)} & \small{1/4(366)} & \small{1/8(183)} & \small{1/16(92)} & 1.4k(1464) &  1/2(5291) & 1/4(2646) & 1/8(1323) & 1/16(662) \\ \midrule
        MT~\cite{tarvainen2017:MT} & 69.16 & 63.01 & 55.81 & 48.70 & - & 77.61 & 76.62 & 73.20 & 70.59 \\
        VAT~\cite{VAT} & 63.34 & 56.88 & 49.35 & 36.92 & - & - & - & - & - \\
        AdvSemSeg~\cite{hung2018adversarial} & 65.27 & 59.97 & 47.58 & 39.69 & 68.40 & - & - & - & - \\
        CCT~\cite{ouali2020CCT} & 62.10 & 58.80 & 47.60 & 33.10 & 69.40 & 77.56 & 76.17 & 73.00 & 67.94 \\
        GCT~\cite{GCT} & 70.67 & 64.71 & 54.98 & 46.04 & - & 77.14 & 75.25 & 73.30 & 69.77 \\
        CutMixSeg~\cite{french2020:cutmixseg} & 69.84 & 68.36 & 63.20 & 55.58 & - & 75.89 & 74.25 & 72.69 & $72.56$ \\ 
        PseudoSeg~\cite{zou2021pseudoseg} & 72.41 & 69.14 & 65.50 & 57.60 & 73.23 & - & - & - & - \\ 
        CPS~\cite{chen2021:CPS} & 75.88 & 71.71 & 67.42 & 64.07 & - & 78.64 & 77.68 & 76.44 & 74.48 \\ 
        PC$^2$Seg~\cite{zhong2021:pixel_contrastive} & 73.05 & 69.78 & 66.28 & 57.00 & 74.15 & - & - & - & - \\  \midrule
        Supervised baseline & 67.67 & 60.63 & 53.03 & 40.31 & 73.71 & 76.02 & 75.23 & 72.87 & 67.12 \\
        RC$^2$L(ours) & \textbf{77.06} & \textbf{72.24} & \textbf{68.87} & \textbf{65.33} & \textbf{79.33} & \textbf{80.43} & \textbf{79.71} & \textbf{77.49} & \textbf{75.56} \\ \bottomrule
    \end{tabular}
    \caption{Comparison with the state-of-the-art methods on VOC 2012 \texttt{Val} set. We use labeled data under all partition protocols to train an original MaskFormer as the supervised baseline. Previous works~\protect\cite{ouali2020CCT,GCT} used the segmentation model which is pretrained on COCO~\protect\cite{COCO} dataset for all partition protocols. We only take COCO pretrained model on 1/4, 1/8 and 1/16 VOC \texttt{Train}, and 1/16 VOC \texttt{Aug}. For the rest of partition protocols, we use the backbone pretrained on ImageNet~\protect\cite{deng2009imagenet}, and initialize the weight of MaskFormer head randomly. All methods are based on ResNet101 backbone.}
\label{tab1}
\end{table*}

\begin{table}[tb]\small
    \centering\setlength{\tabcolsep}{6pt}
    \begin{tabular}{l|c|c}
        \toprule
        Method & 1/4(744) & 1/8(372) \\ \midrule
        AdvSemSeg~\cite{hung2018adversarial} & 62.3 & 58.8  \\
        CutMixSeg~\cite{french2020:cutmixseg} & 68.33 & 65.82 \\ 
        CMB~\cite{alonso2021cmb} & 65.9 & 64.4 \\
        DCL~\cite{lai2021semi} & 72.7 & 69.7 \\
        PseudoSeg~\cite{zou2021pseudoseg} & 72.36 & 69.81 \\
        PC$^2$Seg~\cite{zhong2021:pixel_contrastive} & 75.15 & 72.29 \\  \midrule
        Supervised baseline & 73.94 & 71.53 \\
        RC$^2$L(ours) & \textbf{76.47} & \textbf{74.04} \\ \bottomrule
    \end{tabular}
    \caption{Comparison with the state-of-the-art methods on Cityscapes \texttt{Val} set. We use the model which is pretrained on COCO dataset. All methods are based on ResNet101 backbone.}
    \label{tab2}
\end{table}

\begin{table*}[t]
    \begin{subfigure}[htp]{0.62\linewidth}{
        \fontsize{9}{9}\selectfont 
        \setlength{\tabcolsep}{8pt}{
        \begin{tabular}{c|c|c|c|c|c}
        \toprule
            $\mathcal{L}_{SMC}$ & $\mathcal{L}_{RCC}$ & $\mathcal{L}_{RMC}$  & $\mathcal{L}_{RFC}$ & VOC \texttt{Train} (1/2) & VOC \texttt{Aug} (1/4) \\ \midrule
              ~   &    ~   &    ~   &    ~   & 69.37 & 75.26 \\
            \cmark &    ~   &    ~   &    ~   & 73.26 & 76.58 \\ 
            \cmark & \cmark &    ~   &    ~   & 76.07 & 77.89 \\ 
            \cmark & \cmark & \cmark &    ~   & 76.85 & 78.92 \\ 
            \cmark & \cmark & \cmark & \cmark & \textbf{77.06} & \textbf{79.71} \\ \bottomrule
    \end{tabular}
    \caption{\label{tab3(a)}}}}
    \end{subfigure}
    \hspace{0.27cm}
    \begin{subfigure}[htp]{0.36\linewidth}{
        \fontsize{9}{9}\selectfont 
        \setlength{\tabcolsep}{7pt}{
        \begin{tabular}{c|c|c}
        \toprule
            $\alpha $ & VOC \texttt{Train} (1/2) & VOC \texttt{Aug} (1/4) \\ \midrule
            1.0 & 75.65 & \textbf{79.71} \\
            1.5 & 76.60 & 79.15 \\ 
            2.0 & \textbf{77.06} & 79.15 \\ \bottomrule
        \end{tabular}
        \caption{\label{tab3(b)}}}}
    \end{subfigure}
    \caption{(a) Ablation study of each component. The first row is the result of our baseline model. (b) Effect of different unsupervised loss weights.}
\end{table*}

\section{Experiment}
\subsection{Datasets}
\paragraph{PASCAL VOC 2012.} PASCAL VOC 2012~\cite{everingham2015pascal} is a standard object-centric semantic segmentation dataset, which contains more than 13,000 images with 21 classes (20 object classes and 1 background class).
In the original PASCAL VOC 2012 dataset (VOC \texttt{Train}), 1464 images are used for training, 1449 images for validation and 1456 images for testing. Following the common practice, we also use the augmented dataset (VOC \texttt{Aug}) which contains 10,582 images as the training set. For both the original and augmented datasets, 1/2, 1/4, 1/8, and 1/16 training images are used as labeled data, respectively, for conducting the semi-supervised experiments.

\paragraph{Cityscapes.} Cityscapes~\cite{cordts2016cityscapes} dataset is designed for urban scene understanding with a high resolution ($2048\times1024$). It contains 19 classes for scene semantic segmentation. The finely annotated 5,000 images that follow the official split have 2,975 images for training, 500 images for evaluation, and 1,525 images for testing. We employ 1/4 and 1/8 training images as the labeled data, and the remaining images are used as the unlabeled data.

\paragraph{Evaluation metrics.} We use single scale testing, and report the evaluation results of the model on Pascal VOC 2012 \texttt{val} set and Cityscapes \texttt{val} set by using mean of Intersection over Union (mIoU). 
We compare with state-of-the-art methods on different dataset partition protocols. 

\paragraph{Implementation details.} We use ResNet101 as our backbone, and use MaskFormer Head~\cite{cheng2021:maskformer} as segmentation head. We follow the original MaskFormer to set the hyper-parameters. For all experiments, we set the batch size to 16, and use ADAMW as the optimizer with an initial learning rate of 0.0001, and weight decay of 0.0001. Empirically, we set the loss weight of $\beta_1$, $\beta_2$, $\beta_3$ and $\beta_4$ to 1, 20, 4 and 4, respectively. In addition, we employ a poly learning rate policy which is multiplied by $(1 - \frac{iter}{max_iter})^{power} $ with $power = 0.9$. For PASCAL VOC 2012 dataset, we use the crop size of $512\times512$, and train our model for 120K and 160K iterations for VOC \texttt{Train} and VOC \texttt{Aug}, respectively. For Cityscapes dataset, we use the crop size of $768\times768$, and set training iterations as 120K without using any extra training data.

\subsection{Comparison to the State-of-the-Arts}

\paragraph{Results on PASCAL VOC 2012.} We show the comparison results on the PASCAL VOC 2012 \texttt{val} set in Table \ref{tab1}. All the models are trained using 4 V100 GPUs. On VOC \texttt{Train}, our proposed RC$^2$L outperforms all existing pixel-level regularization methods, achieving the improvement of 1.18\%, 0.53\%, 1.45\%, and 1.26\%  with partition protocols of 1/2, 1/4, 1/8, and 1/16 respectively.
We also compare the results of the 1.4k/9k split, our RC$^2$L is +5.18\% higher than the most recent PC$^2$Seg (79.33 vs. 74.15).
On VOC \texttt{Aug}, RC$^2$L also outperforms the previous state-of-the-art methods, obtaining the improvements of 1.79\%, 2.03\%, 1.05\%, and 1.08\% under 1/2, 1/4, 1/8, and 1/16 partitions, respectively.

\paragraph{Results on Cityscapes.} Table \ref{tab2} shows comparison results on Cityscapes \texttt{val} set. All the models are trained using 8 V100 GPUs. The performance of our method is improved by 2.53\% and 2.51\%, compared to the supervised baseline under partition protocols of 1/4 and 1/8, respectively. Our method also outperforms the previous state-of-the-art. For example, RC$^2$L outperforms PC$^2$Seg, 
which is the best pixel-level contrastive learning method by 1.32\% and 1.75\% under 1/4 and 1/8 partition protocols, respectively.

\subsection{Ablation study}
\paragraph{Investigating each component.} 
We investigate the effect of each component in our methods. The experimental results are illustrated in Table \ref{tab3(a)}.
The baseline model only uses mask consistency between teacher and student predictions for unlabeled data.
It can be seen that the baseline method achieves 69.37\% and 75.26\% on 1/2 VOC \texttt{Train} and 1/4 VOC \texttt{Aug} datasets, respectively. We can see that SMC loss improves the baseline by 3.89\% and 1.32\% on 1/2 VOC \texttt{Train} and 1/4 VOC \texttt{Aug} datasets, respectively. RCC loss obtains the improvement of 2.81\% and 1.31\%  over ``Baseline + SMC loss". These experimental results demonstrate the effectiveness of our region-level consistency learning. Furthermore, RMC loss obtains the improvement of 0.78\% and 1.03\%. RFC loss further achieves the improvement of 0.21\% and 0.79\%. These experimental results demonstrate the effectiveness of our region-level contrastive learning.

\paragraph{Loss weight}
We show the effect of loss weight $\alpha$ which is used to balance the supervised loss and unsupervised loss in Table \ref{tab3(b)}. We find that $\alpha = 2$ achieves the best performance on 1/2 VOC \texttt{Train} set. For 1/4 VOC \texttt{Aug} set, $\alpha = 1$ obtains the best performance.

\begin{table}[t]
    \centering\setlength{\tabcolsep}{7pt}
    \begin{tabular}{c|c|c}
    \toprule
        Num. queries & VOC \texttt{Train} (1/2) & VOC \texttt{Aug} (1/4) \\ \midrule
        100 & 74.01 & 75.83 \\
        50 & \textbf{77.06} & \textbf{79.71} \\ 
        20 & 76.18 & 78.37 \\ \bottomrule
    \end{tabular}
    \caption{Study on the number of queries.}
    \label{tab6}
\end{table}

\paragraph{Number of queries}
We study the number of queries on PASCAL VOC 2012 in Table \ref{tab6}. We can see that $N = 50$ achieves the best result, and $N = 100$ degrades the performance more than $N = 20$. This suggests that a small number of queries is sufficient to provide a good result for datasets with a few number of categories in just one image.

\begin{figure}[t]
\centering
\includegraphics[scale=1.22]{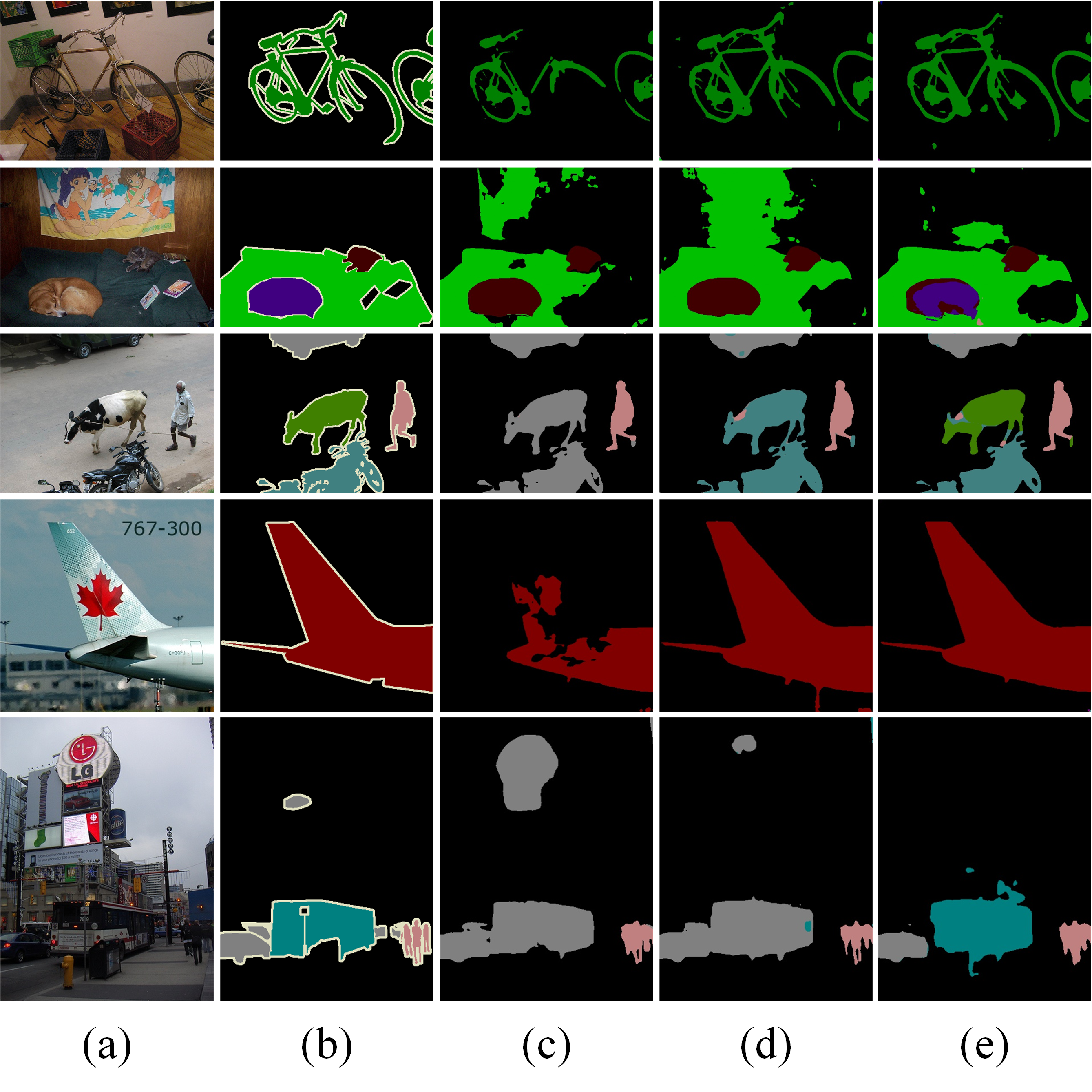}
\caption{Prediction results on PASCAL VOC 2012 \texttt{Val} set under 1/2 (732) partition. (a) original image, (b) ground truth, (c) supervised baseline, (d) semi-supervised consistency baseline, (e) ours.}
\label{fig4}
\end{figure}

\subsection{Visualization}
Figure \ref{fig4} shows the visualization results of our methods on PASCAL VOC 2012. 
We compare the predictions of our RC$^2$L with the ground-truth, supervised baseline, and semi-supervised consistency baseline. 
One can see that our RC$^2$L can correct more noisy predictions compared to the supervised baseline and the semi-supervised consistency baseline. We follow CCT and directly complete consistency learning between student outputs and pseudo labels as the semi-supervised consistency baseline.
In particular, the supervised baseline mislabels some pixels in the 1st row and the 3rd row. Both the supervised baseline and the semi-supervised consistency baseline mistakenly classify some pixels in the 3rd row and the 5th row.

\section{Conclusion}

We have developed the Region-level Contrastive and Consistency Learning (RC$^2$L) method for semi-supervised semantic segmentation. The core contributions of our RC$^2$L are the proposed region-level contrastive and consistency regularization. The former consists of Region Mask Contrastive (RMC) loss and Region Feature Contrastive (RFC) loss, the latter contains Region Class Consistency (RCC) loss and Semantic Mask Consistency (SMC) loss. Extensive experiments on PASCAL VOC 2012 and Cityscapes have shown
that our RC$^2$L outperforms the state-of-the-art semi-supervised semantic segmentation methods, demonstrating that our region-level Contrastive and Consistency regularization can achieve better results than previous pixel-level regularization.

\section*{Acknowledgments}
This work is funded by the Provincial Science and Technology Innovation Special Fund Project of Jilin Province (20190302026GX), Natural Science Foundation of Jilin Province (20200201037JC).

\newpage
\bibliographystyle{named}\small
\bibliography{ijcai22} 

\newpage
\appendix

In the appendix, we first present ablation experiments about temperature hyper-parameter. Then, we provide more visualization results and analysis.

\section{Ablation Experiments}
\paragraph{Temperature hyper-parameter $\tau$}
Table \ref{taba1} ablates different temperature hyper-parameters $\tau_m$ and $\tau_f$ in $\mathcal{L}_{RMC}$ and $\mathcal{L}_{RFC}$. 
We can see that $\tau_m$ = 1.0 and $\tau_f$ = 0.5 achieve the best performance. Different from NT-Xent loss~\cite{chen2020:simclr}, a smaller temperature in $\mathcal{L}_{RMC}$ leads to lower results.

\begin{table}[h]
    \centering\setlength{\tabcolsep}{10pt}
    \begin{tabular}{c|c|c|c}
    \toprule
        $\tau_m$ & $\tau_f $ & VOC \texttt{Train} (1/2) & VOC \texttt{Aug} (1/4)  \\ \midrule
        0.5 & 0.5 & 73.35 & 77.33  \\
        0.5 & 1.0 & 72.39 & 75.69  \\
        1.0 & 0.5 & \textbf{77.06} & \textbf{79.71} \\ 
        1.0 & 1.0 & 74.29 & 78.94 \\ \bottomrule
    \end{tabular}
    \caption{Effect of different temperature hyper-parameters.}
    \label{taba1}
\end{table}

\section{Visualization and Analysis}
\begin{figure*}[t]
\centering
\includegraphics[scale=1.7]{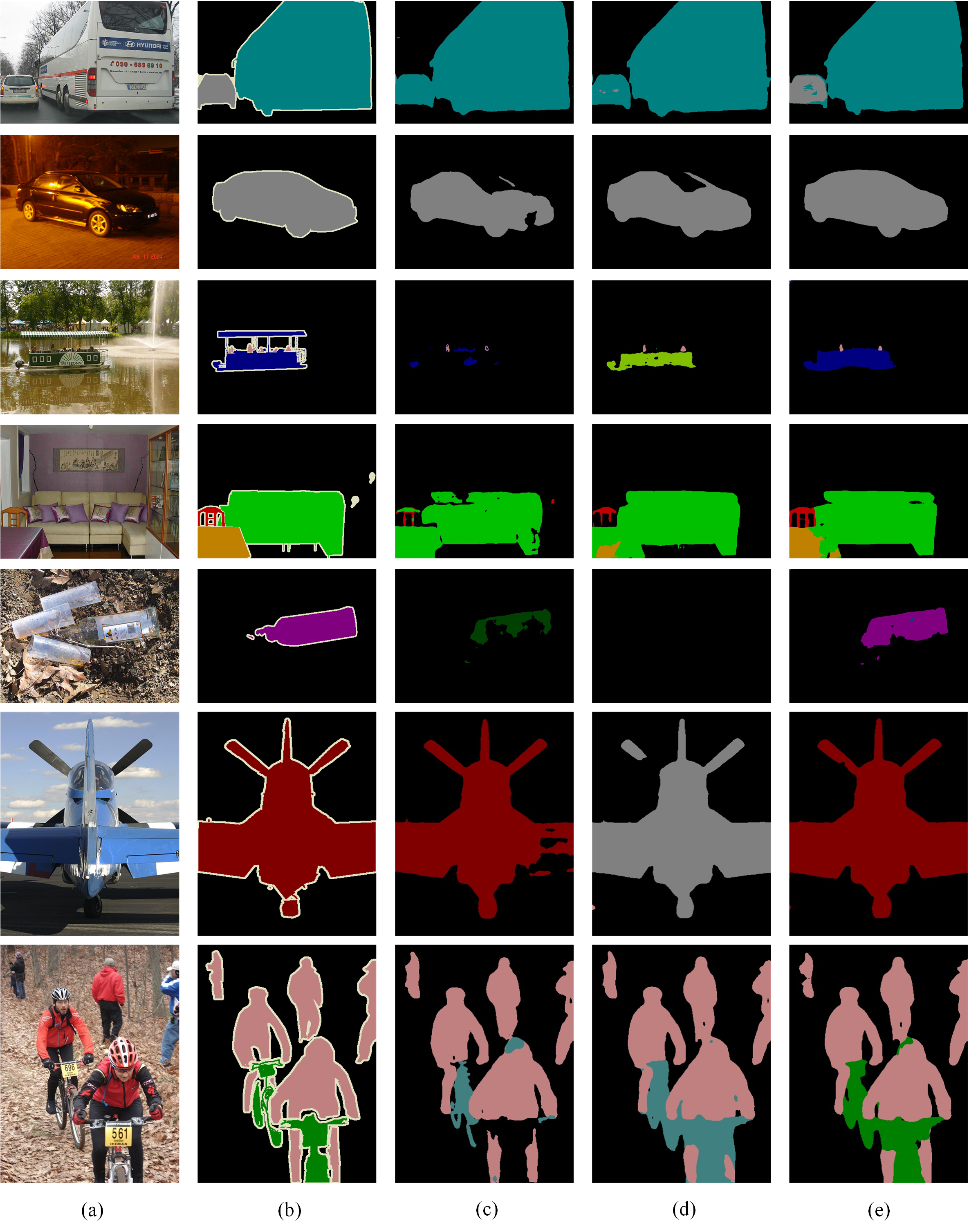}
\caption{Prediction results on PASCAL VOC 2012 \texttt{Val} set under 1/2 VOC \texttt{Train}. (a) original image, (b) ground truth, (c) supervised baseline, (d) semi-supervised consistency baseline, (e) ours.}
\label{figb1}
\end{figure*}

As shown in Figure \ref{figb1}, we further compare the visualization results of our RC$^2$L, supervised baseline, and semi-supervised consistency baseline. We do not compare RC$^2$L with the state-of-the-art methods because their pretrained models are not publicly. It can be seen that the supervised baseline mislabels many pixels due to limited labeled data. For example, in the 3rd row, the supervised baseline mislabels many boat pixels. Supervised and semi-supervised consistency baselines mistakenly classify some regions or pixels. In the 1st row and the 7th row,  both supervised baseline and semi-supervised consistency baseline mistakenly classify the car pixels and bicycle pixels to bus pixels and motorbike pixels, respectively. In particular, we find that semi-supervised consistency baseline has relatively good region mask predictions but mistakenly classifies all the region pixels in the 3rd row, 6th row and 7th row. However, our RC$^2$L can produce significantly less noisy or wrong predictions than the two baselines, which demonstrates the effectiveness of our region-level contrastive and consistency regularization.

\end{document}